\documentclass[11pt,a4paper]{article}
\usepackage[hyperref]{eacl2021}
\usepackage{times}
\usepackage{latexsym}

\usepackage{comment}
\usepackage{microtype}
\usepackage{graphicx}
\aclfinalcopy 

\usepackage[utf8]{inputenc}
\usepackage[arabic,USenglish]{babel}

\usepackage{float}
\restylefloat{table}

\title{TEET! Tunisian Dataset for Toxic Speech Detection}

\author{Slim Gharbi \\ {\bf Heger Arfaoui} \\{Mediterranean Institute of Technology, Tunisia} \\ \texttt{\{slim.gharbi,heger.arfaoui\}@medtech.tn} \\ {\bf Hatem Haddad } \\ {\bf Mayssa Kchaou} \\ { iCompass, Tunisia} \\ \texttt{hatem@icompass.digital, mayssakchaou933@gmail.com }}

\date{}

\begin{document}
\maketitle

\begin{abstract}
The complete freedom of expression in social media has its costs especially in spreading harmful and abusive content that may induce people to act accordingly. Therefore, the need of detecting automatically such a content becomes an urgent task that will help and enhance the efficiency in limiting this toxic spread. Compared to other Arabic dialects which are mostly based on MSA, the Tunisian dialect is a combination of many other languages like MSA, Tamazight, Italian and French. Because of its rich language, dealing with NLP problems can be challenging due to the lack of large annotated datasets. In this paper we are introducing a new annotated dataset composed of approximately 10k of comments. We provide an in-depth exploration of its vocabulary through feature engineering approaches as well as the results of the classification performance of machine learning classifiers like NB and SVM and deep learning models such as ARBERT,     MARBERT and XLM-R.
\end{abstract}

\section{Introduction}
The tunisian dialect or "Derja" is a combination of many languages like MSA, Tamazight, Italian and French ~\cite{fourati}. Studies of tunisian abusive and hate speech detection is still very limited compared to Indo-European languages ~\cite{Haddad}. Studies have been conducted recently addressing Arabic related hate speech detection. For this purpose, Arabic pretrained models like ARBERT and MARBERT (Abdul-Mageed et al., 2020) have been introduced and have shown their ability in achieving different NLP tasks especially sentiment analysis in Arabic context. Another kind of deep learning model, called XLM-R, has been introduced by Facebook with its characteristic of being pretrained on multiple languages has achieved outstanding results while dealing with underrepresented languages (Conneau et al.,2020). The paper is structured as follows: Section 2 illustrates the previous works made on detecting hate speech in tunisian dialect. In section 3, we are defining the methodology that has been followed for each process of achieving this task. Followed by section 4, where we introduce the results obtained from the conducted experiments. Section 5 provides the limitation of our work. Finally, section 6 concludes by stating the contributions of this study as well as possible future works.

\section{Related Works}
In our context of detecting hate and abusive speech for tunisian dialect, the only existing annotated dataset is T-HSAB combining 6,039 annotated comments as hateful, abusive or normal. ~\cite{Haddad}. Besides introducing the annotation of the collected dataset and its evaluation, the authors have experimented as well two machine learning algorithms: SVM and NB. They have benchmarked the performance of these classifiers alongside different feature engineering approaches and classification tasks. The best results obtained have been achieved while discriminating between only 2 classes, using uni + bi grams and Term frequency {$ \geq 2$} with F1-score of 92.3 for NB and 83.6 for SVM. None of the deep learning models have been experimented on the introduced dataset.
\section{Methodology}
\subsection{Data}
\subsubsection{Data Collection}
One of the main reasons that induces our choice of collecting a new dataset is the lack of annotated datasets for such underrepresented languages like tunisian. Most importantly, the unique available dataset combines only 6039 labeled comments with exactly 3834 classified as normal. This distribution of data shows that T-HSAB is imbalanced towards inoffensive language. Therefore, the need of collecting data with abusive connotation has been emerged. 
The authors of T-HSAB have the idea to identify keywords that lead to abusive or hate classification. This identified list has been used in order to target and extract tunisian toxic comments from different social media platforms by injecting queries containing these specific words.
\subsubsection{Data Annotation}
Before proposing any annotation guidelines, we have relied on the literature and have been inspired from different annotation guidelines suggested for Arabic hate and abusive datasets. In this specific task, we have faced a trade-off between data usability and data investigation. To make more clear the difference between these two strategies: There are some annotators who go for straightforward schemes by defining only one layer of annotation either binary or 3-way classification with the purpose of getting an annotated dataset in less time to start the modeling part. In the other hand, some annotators want to dig more deeply in the problem investigation in order to understand the phenomena. In this way, they are choosing more sophisticated schemes like adding a second layer of classification and differentiating between the intention of the comment's writer whether it is implicit or explicit. For example, a toxic comment may be disguised with sarcasm and therefore considered as implicit. This kind of implicitness is hard to detect especially for tunisian dialect. For all these reasons, we have relied on the simplest and the mostly used annotation scheme which is a 3-way classification: Hate, Abusive and Normal. 

\subsubsection{Feature Engineering}
Feature engineering is an essential part before switching to the machine learning modeling. The purpose is to represent and transform raw data into insightful data. In this section, we have applied and experimented different n-gram schemes in addition to removing the stopwords as well as limiting the number of features.
We have used Bag of Words (BoW) for the representation of our data aligning each term with its frequency.
\begin{itemize}
    \item Removal of Stopwords: After investigating the vocabulary in our dataset and looking for the most frequent words, we have identified a list of stopwords that do not add any contextual meaning.
   
   \item N-gram scheme: We have taken into consideration single words (Uni-gram), sequences of two successive words (Uni + Bi grams) as well as expressions of three successive words (Uni + Bi + Tri).
   
   \item Reduce the feature size: Only words or terms with frequencies {$\geq 2$} have been considered. We have assumed that terms that appear only once in our dataset do not have any impact in the understanding of our corpus.
   
   \item Data Distribution Imbalance: We have to choose between three approaches:
   \begin{itemize}
       \item Undersampling the majority represented classes by getting rid of a number of comments in order to have a number of comments that is near to the minority class.
       \item Oversampling by replicating comments related to minority classes such that we are targeting a number of comments that is close to the majority class.
       \item Dividing the comments that are of high length into multiple comments with the same label.
   \end{itemize}
   
\end{itemize}

\subsection{Model}
\subsubsection{Machine Learning}
For machine learning model evaluation, we use scikit-learn package and experiment different combination of feature engineering techniques alongside different algorithms: MultinomialNB, SVM, Random Forest and Multinomial Logistic Regression. The metrics used to evaluate the model's predictions are the accuracy, recall, precision and F1-Score. The latter has been the decisive metric used for comparison. The data has been split into 0.8 for training and 0.2 for testing. Experiments have been conducted on different datasets:
\begin{itemize}
    \item 1st dataset: The new collected dataset combining 10091 labeled comments.
    \item 2nd dataset: The merge between the new collected dataset and T-HSAB combining 16130 annotated comments.
\end{itemize}

\subsubsection{Deep Learning}
We have decided to use the new architecture of transformers. Our choice has been based on their ability of treating data in parallel compared to the CNNs and RNNs. For this purpose, we have the idea to experiment different type of transforms that follow the same architecture of BERT-Base. 
\begin{itemize}
    \item Monolingual Models: Models that are pre-trained only on a single language: MARBERT and ARBERT
    \item Multilingual Model: Model that is pre-trained on multiple languages: XLM-R
\end{itemize}
Since building a model from scratch is costly in terms of time and resources, we have decided to apply the transfer learning by freezing all the architecture of the model and only make the changes on the input layer and the output layer depending on the classification task that we want to evaluate.
Only the merged dataset presented previously has been used for the deep learning classification. The only difference is none of the feature engineering techniques discussed in previous section have been applied. We have conducted 12 experiments for each model defined in the tab.\ref{tab:exphyperparemters}

\begin{table*}[htbp]
\begin{tabular}{ |p{3cm}||p{3cm}||p{3cm}||p{4cm}| }
 \hline
 \multicolumn{4}{|c|}{Hyper-parameters} \\
 \hline
 Experiment Number   & Batch Size  & Learning Rate & Number of Epochs\\
 \hline
 \hline
 1 & 16 & 5e-5 & 3\\
 \hline
 \hline
 2 & 16 & 5e-5 & 4\\
 \hline
 \hline
 3 & 16 & 3e-5 & 3\\
 \hline
 \hline
 4 & 16 & 3e-5 & 4\\
 \hline
 \hline
 5 & 16 & 2e-5 & 3\\
 \hline
 \hline
 6 & 16 & 2e-5 & 4\\
 \hline
 \hline
 7 & 32 & 5e-5 & 3\\
 \hline
 \hline
 8 & 32 & 5e-5 & 4\\
 \hline
 \hline
 9 & 32 & 3e-5 & 3\\
 \hline
 \hline
 10 & 32 & 3e-5 & 4\\
 \hline
 \hline
 11 & 32 & 2e-5 & 3\\
 \hline
 \hline
 12 & 32 & 2e-5 & 4\\
 \hline
 \hline 
\end{tabular}
\centering
\caption{\label{tab:exphyperparemters}Experiments of Hyper-parameters.}
\end{table*}

\section{Results and Analysis}
\subsection{Annotation}
In the table.\ref{tab:1} below, we have defined the classification scheme:
\begin{table*}[h!]
\begin{tabular}{ |p{3cm}||p{12cm}|  }
 \hline
 \textbf{Classification}     & \textbf{Annotation Guidelines} \\
 \hline
 Hate   & Explicit or implicit hostile intention that threats or projects
violence,targeting a group of individuals based on common specifications
like race, ethnicity, gender, religion, political ideologies\\
 \hline
 \hline
 Abusive   & Instances that contain offensive content, profanity, vulgarity by mentioning
private physical part or sexual acts, offensive content or insulting
individuals based on physical or mental characteristics using animal
analogies , wishing evil\\
 \hline
 \hline
 Normal   & --\\
 \hline
\end{tabular}
\centering
\caption{\label{tab:1}Annotation Guidelines}
\end{table*}

In addition, due to the fact that the boundaries between hate and offensive comments are quite thin, we have added some extra examples to state the differences. For example if the offense is ethnic related, it is directly categorized as hate comment. But if we have an offense against gender especially women, we have to look if there is a sexual connotation in the comment and if the target is an individual or a group to figure out to which category it relates.  If it contains vulgar content and directed towards one individual than it is an abusive comment otherwise it relates to hate comment. Example: She is a bitch is an abusive comment vs. all women are bitches is a hate comment.

Furthermore, for experimental purposes, we have defined a 2nd layer of annotation related to class hate with sub-categories: Racism, Sexism, Homophobic, Religion and Others. In class others, we have considered hate speech related to political affiliation, regionalism etc.

When it comes to the annotation, two native tunisian speakers shared the task of annotating the collected dataset by splitting it by two.

According to the annotation guidelines and definitions, the comments have been labeled. Examples are shown in the table.\ref{tab:55}
\begin{table*}[h!]
\begin{tabular}{ |p{3cm}||p{12cm}|  }
 \hline
 \textbf{Class}     & \textbf{Comments} \\
 \hline
 Hate   & \textRL{\foreignlanguage{arabic}{تونس يا بلد العهر}} (Tunisia country of immorality)\\
 \hline
 \hline
 Abusive   & \textRL{\foreignlanguage{arabic}{هاذي مرا رخيصة و متسواش}} (This woman is despicable and worthless)\\
 \hline
 \hline
 Normal   & \textRL{\foreignlanguage{arabic}{ولي معندهاش ميزانية منين}} (and the one who does not have the budget) \\
 \hline
\end{tabular}
\centering
\caption{\label{tab:55}Examples of Annotated Comments}
\end{table*}

We have made some exploratory analysis to understand more our data. First we have to look after the distribution of the classes. As shown in the table.\ref{tab:distribution of the new data} below, most of our data is classified as abusive with more than 6K out of 10k. The class abusive with all its categories is highly represented with more than 3k of comments. Finally, the class normal is low represented with less than 1k. It is consistent with the extraction and collection of data strategy, since we have targeted and filtered comments towards abusive and hate content. When it comes to categories related to hate label, the others category is the most represented one. With nearly the same number of comments, the category racism comes right after. Sexism is represented with only 642 comments out of 3155 hate comments. Both homophobic and religious hate speech classes are of very low distribution.

\begin{table}[h!]
\begin{tabular}{ |p{3cm}||p{3cm}|  }
 \hline
 \multicolumn{2}{|c|}{Data Distribution} \\
 \hline
 Label & Number of Comments\\
 \hline
 \hline
 -1 & 6118\\
 \hline
  \hline
 -2 & 3155\\
 \hline
  \hline
 0 & 818\\
 \hline
  \hline
 -2 (Others) & 1179\\
 \hline
  \hline
 -2 (Racism) & 1027\\
 \hline
  \hline
 -2 (Sexism) & 642\\
 \hline
  \hline
 -2 (Homophobic) & 173\\
 \hline
  \hline
 -2 (Religion) & 134\\
 \hline
 \hline
\end{tabular}
\centering
\caption{\label{tab:distribution of the new data}Distribution of the Dataset }
\end{table}

Since we are dealing with text and word tokenization, we have investigated the most frequent words in our dataset. Some examples of the 20 most frequent words are illustrated in the tab.\ref{tab:most frequent words} below. The uni-gram words are mostly represented by Arabic stopwords. They are relevant in tunisian dialect and are expected to appear frequently. In addition, we remark the presence of the most abusive words used in daily life. Finally there are words that are related to specific subjects like ”God” when invoking religious subjects or when swearing.

\begin{table}[h!]
\begin{tabular}{ |p{3cm}||p{3cm}|  }
 \hline
 \multicolumn{2}{|c|}{Most Frequent Words} \\
 \hline
 Word & Word Frequency\\
 \hline
 \hline
 \textRL{\foreignlanguage{arabic}{ و}}(and) & 7091\\
 \hline
 \hline
 \textRL{\foreignlanguage{arabic}{ في }}(in) & 4140\\
 \hline
 \hline
 \textRL{\foreignlanguage{arabic}{ من }} (from) & 2732\\
 \hline
 \hline
 \textRL{\foreignlanguage{arabic}{ يا }} (oh) & 2219\\
 \hline
 \hline
 \textRL{\foreignlanguage{arabic}{ على}}(on) & 1949\\
 \hline
 \hline
 \textRL{\foreignlanguage{arabic}{تونس}} (Tunisia) & 1487\\
 \hline
 \hline
 \textRL{\foreignlanguage{arabic}{ الله }} (God) & 910\\
 \hline
 \hline
 \textRL{\foreignlanguage{arabic}{ كل }}(all) & 709\\
 \hline
 \hline
 \textRL{\foreignlanguage{arabic}{ طحان }} (Profane Tunisian Language Word) & 709\\
 \hline
 \hline
 \textRL{\foreignlanguage{arabic}{ هذا }}(this) & 596\\
 \hline
 \hline
\end{tabular}
\centering
\caption{\label{tab:most frequent words}Most Frequent Words }
\end{table}

\begin{figure}[htbp]
\centerline{\includegraphics[width=9cm]{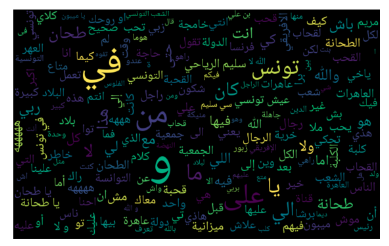}}
\caption{Dataset WordCloud.}
\label{fig:Wordcloud new dataset}
\end{figure}
Fig \ref{fig:Wordcloud new dataset} shows as well the most frequent words in our dataset. \textRL{\foreignlanguage{arabic}{ تونس}} ("Tunisia") is highly invoked within abusive language for example and this is justifying that entities are targeted while using profane language.

\subsection{Feature Engineering}
\begin{figure}[htbp]
\centerline{\includegraphics[scale=.5]{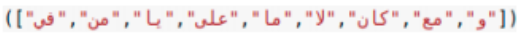}}
\caption{List of Identified Arabic Stopwords}
\label{fig:arabic stopwords}
\end{figure}

After exploring our dataset and the most frequent words, we have identified the following list of stopwords shown in fig.\ref{fig:arabic stopwords}.
We have to mention that this list is error prone because it relies on a subjective interpretation.
In the table. \ref{tab:most frequent words}, we have illustrated highly represented words like [ \textRL{\foreignlanguage{arabic}{ هذا}} \textRL{\foreignlanguage{arabic}{ كل }}]. We have chosen to not remove these specific words because they are pointing either one specific person or a group of people. This criteria of differentiating between the type of target is crucial when discriminating between hate and abusive classes. 

One of the mentioned feature engineering approach has been the application of different N-gram schemes. When it comes to bi-gram expressions, we remark the presence of a political party "\textRL{\foreignlanguage{arabic}{عيش تونسي}}"(Aïch Tounsi) appearing 169 times, the name of politicians like the tunisian ex-president "\textRL{\foreignlanguage{arabic}{ بن علي}}" (Ben Ali) and the president of a political party "\textRL{\foreignlanguage{arabic}{ سليم رياحي }}"(Slim Riahi) with frequencies of 88 and 92 respectively, a soccer club "\textRL{\foreignlanguage{arabic}{ النادي الإفريقي}}"( African Club) 
showing 49 times or a tunisian celebrity "\textRL{\foreignlanguage{arabic}{  مريم الدباغ }}"(Meriem Debbagh) appearning 68 times as well as words generalizing all the Tunisians "\textRL{\foreignlanguage{arabic}{  الشعب التونسي }}" (Tunisian People) with a frequency of 84. Of course there are many profane tunisian language identified in the bi-grams.
Regarding the tri-grams, we have remarked expressions containing replicated words showing the insistence of the writer. We have ended up with getting rid of the tri-grams since they do not add any contextual meaning to our corpus.

Since we are choosing words with term frequency {$\geq 2$}, we can have a ranking of all the words based on their frequency in a csv file and visualize the number of max-features to be taken for each n-gram scheme. The results are represented in the table \ref{tab:Number of max features} below:
\begin{table*}[h!]
\begin{tabular}{ |p{3cm}||p{3cm}||p{3cm}| }
 \hline
 \multicolumn{3}{|c|}{Number of max-features} \\
 \hline
 N-gram Scheme & Number of features & Number of features {$\geq  2$}\\
 \hline
 \hline
 Uni & 49973 & 17633\\
 \hline
  \hline
 Uni + Bi & 208222 & 31834\\
 \hline
  \hline
 Uni + Bi + Tri & 380008 & 37551\\
 \hline
 \hline
\end{tabular}
\centering
\caption{\label{tab:Number of max features}Number of max features for each N-gram Scheme }
\end{table*}

\subsection{Machine Learning}
In the machine learning section, we have experimented different classification tasks with different datasets. We will divide the obtained results depending on the dataset that has been used.
\paragraph{Experiments on the new collected dataset}
\subparagraph{Experiment 1: 3-way Classification}
Since abusive class is the class with the highest distribution, we oversample the 2 other minority classes hate and normal to get a balanced data. (Number of abusive comments has been divided by 1.5). Results are shown in table.\ref{tab:train oversampled unigram new data}  
\begin{table*}[h!]
\begin{tabular}{ |p{3cm}||p{3cm}||p{3cm}| }
 \hline
 \multicolumn{3}{|c|}{Training Set Oversampled} \\
 \hline
 Class & Training Set & Test Set \\
 \hline
 \hline
 Abusive (-1) & 4910 & 1208 \\
 \hline
  \hline
 Hate (-2) & 3273 & 658 \\
 \hline
  \hline
 Normal (0) & 3273 & 153 \\
 \hline
 \hline
\end{tabular}
\centering
\caption{\label{tab:train oversampled unigram new data}Training Set Oversampled for the new Dataset (3-way) }
\end{table*}

The results illustrated in the table.\ref{tab:MLR3way} below represents the performance of ML classifiers in the 3-way classification for each n-gram scheme.
\begin{table*}[h!]
\begin{tabular}{ |p{5cm}||p{3cm}||p{3cm}||p{3cm}| }
 \hline
 \multicolumn{4}{|c|}{F1-Score} \\
 \hline
 Classifier/N-gram   & Uni  & Uni + Bi & Uni + Bi + Tri\\
 \hline
 \hline
 NB & 0.81 & 0.79 & 0.76\\
 \hline
  \hline
 SVM & 0.86 & 0.86 & 0.86\\
 \hline
  \hline
 Random Forest & \textbf{0.89} & \textbf{0.89} & \textbf{0.89} \\
 \hline
 \hline
 Logistic Regression & 0.87 & 0.88 & 0.87 \\
 \hline
 \hline
\end{tabular}
\centering
\caption{\label{tab:MLR3way}Machine Learning Results on the new Dataset (3-way) }
\end{table*}

The first interpretation is that NB performs better when uni-gram is chosen whereas logistic regression predicts more efficiently when uni+bi gram is applied. The two other models SVM and Random forest keep the same performance whatever the n-gram scheme.
As it can be observed from the table, random forest and logistic regression classifiers are the best performers among all the algorithms (slightly better than SVM). NB is the algorithms with the less efficiency in all n-grams.

\begin{figure}[htbp]
\centerline{\includegraphics[scale=.75]{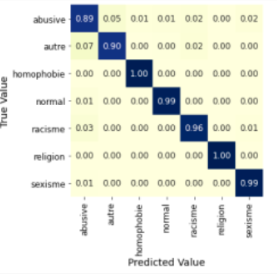}}
\caption{Confusion Matrix uni+bi for 7-way Classification}
\label{fig:7way}
\end{figure}

\subparagraph{Experiment 2: 7-way Classification}
In this experiment, we have considered the 2nd level of classification which consists of the 5 categories related to hate label. In this way, the comments can be classified into 7 categories: Normal, Abusive, Religion, Sexism, Racism, Homophobic and Others.
Since abusive class is the class with the highest distribution, we oversample all the other minority classes to get a balanced data. (Number of abusive comments has been divided by 2).
After splitting data into train (0.8) and test set (0.2) we get the following distribution represented below in table.\ref{tab:traintestsubcategories}:
\begin{table*}[h!]
\begin{tabular}{ |p{3cm}||p{3cm}||p{3cm}||p{3cm}| }
 \hline
 \multicolumn{4}{|c|}{Train Test splitting} \\
 \hline
 Class & Training Set Before Oversampling & Training Set After Oversampling & Test Set \\
 \hline
 \hline
 Abusive (-1) & 4910 & 4910 & 1208\\
 \hline
 \hline
 Others (-2) & 937 & 2455 & 242\\
 \hline
 \hline
 Racism (-2) & 820 & 2455 & 207\\
 \hline
 \hline
 Normal (0) & 665 & 2455 & 153\\
 \hline
 \hline
 Sexism (-2) & 501 & 2455 & 141\\
 \hline
 \hline
 Homophobic (-2) & 129 & 2455 & 44\\
 \hline
 \hline
 Religion (-2) & 110 & 2455 & 24\\
 \hline
 \hline
\end{tabular}
\centering
\caption{\label{tab:traintestsubcategories}Train Test Splitting Distribution of the new Dataset (7-way classification) }
\end{table*}
\newline
The results of F1 scores obtained range from 91 to 96 which represents a high rate. It shows that all the classifiers are delivering a good performance but the fact that the training dataset is mostly composed of replicated comments precisely for low represented classes like religion, homophobic and sexism will result to an overfitting problem. The confusion matrix shown in the fig.\ref{fig:7way} below approves the results of F1-scores. As we can see, we have a classification correctness of 100\% for minority classes like Religion and Homophobic.

\paragraph{Experiments on the merge of T-HSAB and new dataset}
\subparagraph{Experiment 3: 3-way Classification}

Exactly the same procedure has been followed for the experiments related to the merge between the 2 datasets. Our new dataset counts 16115 of annotated comments.
After splitting data into train (0.8) and test set (0.2) we get the following distribution represented below in table.\ref{tab:traintestMergedDistrib}:
\begin{table*}[h!]
\begin{tabular}{ |p{3cm}||p{3cm}||p{3cm}| }
 \hline
 \multicolumn{3}{|c|}{Train Test splitting} \\
 \hline
 Class & Training Set & Test Set\\
 \hline
 \hline
 Abusive (-1) & 5818 & 1426\\
 \hline
  \hline
 Hate (-2) & 3367 & 866\\
 \hline
  \hline
 Normal (0) & 3707 & 931\\
 \hline
 \hline
\end{tabular}
\centering
\caption{\label{tab:traintestMergedDistrib}Train Test Splitting Distribution of the Merged Dataset (3-way classification) }
\end{table*}

Since abusive class is the class with the highest distribution, we oversample the 2 other minority classes hate and normal to get a balanced data. (Number of abusive comments has been divided by 1.2). Results are shown in table.\ref{tab:train oversampled merged}  
\begin{table}[h!]
\begin{tabular}{ |p{3cm}||p{3cm}| }
 \hline
 \multicolumn{2}{|c|}{Training Set Oversampled} \\
 \hline
 Class & Training Set \\
 \hline
 \hline
 Abusive (-1) & 5818 \\
 \hline
  \hline
 Hate (-2) & 4848 \\
 \hline
  \hline
 Normal (0) & 4848 \\
 \hline
 \hline
\end{tabular}
\centering
\caption{\label{tab:train oversampled merged}Training Set Oversampled of the Merged Dataset (3-way Classification)}
\end{table}

We are interested in the F1 score results of the classifiers for each n-gram scheme. Results will be summarized in the table.\ref{tab:RMLMErged}
\begin{table*}[h!]
\begin{tabular}{ |p{5cm}||p{3cm}||p{3cm}||p{3cm}| }
 \hline
 \multicolumn{4}{|c|}{F1-Score} \\
 \hline
 Classifier/N-gram   & Uni  & Uni + Bi & Uni + Bi + Tri\\
 \hline
 \hline
 NB & 0.77 & \textbf{0.8} & 0.79\\
 \hline
  \hline
 SVM & 0.83 & \textbf{0.84} & 0.83\\
 \hline
  \hline
 Random Forest & \textbf{0.84} & 0.83 & \textbf{0.84} \\
 \hline
 \hline
 Logistic Regression & 0.82 & \textbf{0.85} & 0.84 \\
 \hline
 \hline
\end{tabular}
\centering
\caption{\label{tab:RMLMErged}Machine Learning Results on the Merged Dataset (3-way)}
\end{table*}

As it can be observed, the results of F1-scores are ranging from 77 to 85. Random forest outperforms the other classifiers for both uni and uni + bi + tri grams while showing satisfactory results in uni + bi.
Logistic regression is the best classifier with 85\% (uni + bi gram). 
We can conclude also that all the ML models beside random forest, performs better in the uni +bi gram scheme.
Compared to the results obtained in 3-way classification task for the new dataset, we loose some performance even though the number of comments increases from 10 k to 16 k. So the results are not improved by enlarging the dataset. It may be of the annotation guidelines that differ from one dataset to another. Another reason may be the fact that T-HSAB contains a larger number of normal comments than the new dataset. We may suggest to reannotate the datasets using the same annotation guidelines for both datasets. The most important result is that we provide the best F1 score = 85\% in 3-way classification using uni + bi and TF$>$2 in comparison of what have been reported in T-HSAB: 83.6\% for NB.

\subparagraph{Experiment 4: Binary Classification}
In this experiment, we are considering only two classes normal and abusive which has been merged with hate class.
After splitting data into train (0.8) and test set (0.2) we get the following distribution represented below in table.\ref{tab:traintestMergedBinary}:
\begin{table*}[h!]
\begin{tabular}{ |p{3cm}||p{3cm}||p{3cm}| }
 \hline
 \multicolumn{3}{|c|}{Train Test splitting} \\
 \hline
 Class & Training Set & Test Set\\
 \hline
 \hline
 Abusive (-1) & 9185 & 3707\\
 \hline
  \hline
 Normal (0) & 2292 & 931\\
 \hline
 \hline
\end{tabular}
\centering
\caption{\label{tab:traintestMergedBinary}Train Test Splitting Distribution of the Merged Dataset (Binary Classification) }
\end{table*}

Since abusive class is the most represented one, we oversample the class normal to get a balanced data. (Number of abusive comments has been divided by 1.2). Results are shown in table.\ref{tab:train oversampled merged binary}  
\begin{table}[h!]
\begin{tabular}{ |p{3cm}||p{3cm}| }
 \hline
 \multicolumn{2}{|c|}{Training Set Oversampled} \\
 \hline
 Class & Training Set \\
 \hline
 \hline
 Abusive (-1) & 9185 \\
 \hline
  \hline
 Normal (0) & 7654 \\
 \hline
 \hline
\end{tabular}
\centering
\caption{\label{tab:train oversampled merged binary}Training Set Oversampled of the Merged Dataset (Binary Classification)}
\end{table}

The F1 score results of the classifiers for each n-gram scheme are summarized in the table \ref{tab:MLRMergedBinary}
\begin{table*}[h!]
\begin{tabular}{ |p{5cm}||p{3cm}||p{3cm}||p{3cm}| }
 \hline
 \multicolumn{4}{|c|}{F1-Score} \\
 \hline
 Classifier/N-gram   & Uni  & Uni + Bi & Uni + Bi + Tri\\
 \hline
 \hline
 NB & 0.9 & 0.9 & 0.89\\
 \hline
  \hline
 SVM & 0.91 & 0.92 & 0.91\\
 \hline
  \hline
 Random Forest & \textbf{0.93} & 0.92 & 0.92 \\
 \hline
 \hline
 Logistic Regression & 0.92 & \textbf{0.93} & \textbf{0.93} \\
 \hline
 \hline
\end{tabular}
\centering
\caption{\label{tab:MLRMergedBinary}Machine Learning Results on the Merged Dataset (Binary)}
\end{table*}

Most of the F1-scores obtained are above 90\% which induce that our classifiers performs outstandingly.  
For uni-gram, random forest shows the best F1-score with 93\% whereas the Logistic regression shows the same score for Uni+Bi and Uni+Bi+Tri. The best results have been obtained for both schemes uni and uni + bi. T-HSAB has shown a 92.3\% in binary classification as the best performance. In this experiment also, we provide the best result reported in the state of the art.

\subsection{Deep Learning}
We illustrate the results obtained from experimenting monolingual and multilingual deep learning models on our dataset. The used dataset is the merge between T-HSAB and the new collected one in order to satisfy a larger number of comments (16k) which is more appropriate for deep learning application.

The data has been split into training, validation and testing sets. Like stated in previous section, we have tried several training hyperparameters for transfer learning so that we conduct 12 experiments for each model.
\begin{itemize}
    \item Batch Size: 16,32
    \item Learning Rate (Adam Optimizer): 5e-5, 3e-5, 2e-5
    \item Number of Epochs: 3, 4
\end{itemize}
The sequence length was fixed to 128 even though most of our comments are composed of 21 words.
Because of distribution imbalance, we have updated the initial class weights before training our model:
\begin{itemize}
    \item Binary classification: [1.73746631 (abusive) 0.7020257 (normal)] 
    \item 3-Way Classification: [1.15834874 (hate) 0.74147111 (normal) 1.26898414 (abusive)]
\end{itemize}

It is common and  known as a good practice to have a single metric evaluation to compare results and decide which model performs better. Since the classes in our dataset are imbalanced, we rely on the macro F1-Score as evaluation metric for all the experiments. This will allow us also to compare between the outputs of coarse-grained binary classification and fine-grained 3-way classification. Some of the experiments with higher batch size could not be conducted due to resources limitations. This is related to constraints related to the power of GPU provided by Google Colab. To train more sophisticated and heavy models we will need an upgrade to Google Colab pro.  

\paragraph{Monolingual results}
The best MARBERT and ARBERT obtained results for binary and 3-way classification are illustrated in the tables \ref{tab:marbert results} and \ref{tab:arbert results} below:
\begin{table*}[h!]
\begin{tabular}{ |p{2cm}||p{4cm}||p{4cm}| }
 \hline
 \multicolumn{3}{|c|}{F1-Score} \\
 \hline
 Experiment   & Binary Classification  & 3-way Classification\\
 \hline
 \hline
 1 & 0.74 & \textbf{0.66} \\
 \hline
 \hline
 2 & \textbf{0.78} & \textbf{0.66}  \\
 \hline
 \hline
\end{tabular}
\centering
\caption{\label{tab:marbert results}MARBERT Results}
\end{table*}

\begin{table*}[h!]
\begin{tabular}{ |p{2cm}||p{4cm}||p{4cm}| }
 \hline
 \multicolumn{3}{|c|}{F1-Score} \\
 \hline
 Experiment   & Binary Classification  & 3-way Classification\\
 \hline
 \hline
 1 & 0.70 & 0.6 \\
 \hline
 \hline
 2 & \textbf{0.74} & 0.63 \\
 \hline
 \hline
 3 & 0.7 & \textbf{0.64} \\

 \hline
 \hline
\end{tabular}
\centering
\caption{\label{tab:arbert results}ARBERT Results}
\end{table*}

As we notice, for both monolingual transformers, the F1-scores decrease when the classification task switches to 3-way. Another remark is that MARBERT outperforms ARBERT for both classification tasks which confirms what have been stated in the state of the art: Models that are pretrained on arabic dialects perform better than those who are pretrained on only MSA. With 78\% of F1 Score MARBERT delivers its best performance whereas ARBERT shows only 74\% in binary classification.
For experiments 1 and 2, MARBERT performs its best score with 66\%. In the other hand, ARBERT shows 64\% in experiment 3 as its greatest output for the 3 way classification.

\paragraph{Multilingual results}
In the table \ref{tab:xlmr results} below, XLM-R shows outstanding result in experiment 3 when it comes to binary classification with 85\% as F1-Score outperforming the monolingual transformers.
The best results in 3-way classification have been obtained in experiment 1 and 2 with 75\% of F1-Score.
\begin{table*}[h!]
\begin{tabular}{ |p{2cm}||p{4cm}||p{4cm}| }
 \hline
 \multicolumn{3}{|c|}{F1-Score} \\
 \hline
 Experiment   & Binary Classification  & 3-way Classification\\
 \hline
 \hline
 1 & 0.82 & \textbf{0.75} \\
 \hline
 \hline
 2 & 0.83 & \textbf{0.75} \\
 \hline
 \hline
 3 & \textbf{0.85} & 0.74 \\
 \hline
 \hline
\end{tabular}
\centering
\caption{\label{tab:xlmr results}XLM-R Results}
\end{table*}

Based on the results obtained, multilingual transformers are better suited for our context. In other words, for a specific classification task whether binary or 3-way the results have shown that the F1-scores obtained by XLM-R (85\% for binary , 75\% for 3-way) are better than those of ARBERT (74\% for binary , 64\% for 3-way) and MARBERT (78\% for binary , 66\% for 3-way). Therefore, models that are pretrained on multiple languages conclude to better results that those who are pretrained on single languages. Words that do share the same meaning in different languages are mapped to similar vectors. 

All the greatest performances of all models have been obtained in experiment 1,2 and 3. Therefore we will investigate their sharing similarities and differences in relation with the other experiments to distinguish the best training hyper-parameters. In comparison with the other experiments, exp.1,2 and 3 share the same number of batch size = 16. The only differences reside in the learning rate = 5e-5 or 3e-5 and the number of epochs = 3 or 4. Therefore the list of hyperparameters that do have better impact in the classification performance are:
\begin{itemize}
    \item Batch Size = 16
    \item Learning Rate = 5e-5 or 3e-5
    \item Number of Epochs = 3, 4
\end{itemize}

\section{Limitations}
One of the main limitations is that evaluation metrics related to inter-agreement have not been performed on the new annotated dataset. This is due to the fact that the comments have been divided into two chunks and each portion has been delegated to different annotator. Such metrics can provide insights about the reliability of the data.
Another limitation is that no preprocessing or feature engineering tasks have been applied before deep learning modeling and this may be the cause of the underperformance of deep learning classifiers in comparison with machine learning algorithms. 
The evaluation of the performance of deep learning is subject to hyperparameters choices while applying transfer learning. It may be the case that we have not considered and experimented another parameters' values that can conclude to better classification performance results.

\section{Conclusion}
This paper describes a full system of detecting automatically tunisian hate speech from data collection and annotation to the model's application. Various models have been investigated as well as different classification tasks have been experimented. Results have shown that machine learning classifiers outperform deep learning models in our case. Furthermore,the performance of the classifiers has been always better when solving a binary classification problem rather than a 3-way classification. This is justifying that it is easier to discriminate between normal and profane language rather than differentiating between hate and abusive speech due to thin boundaries even in their definitions. As future works, we recommend to involve each annotator in the annotation of the whole dataset in order to evaluate the reliability of the annotated dataset as well as including some preprocessing tasks before deep application. 

\bibliography{anthology,eacl2021}
\bibliographystyle{acl_natbib}

\end{document}